\documentclass[runningheads]{llncs}
\usepackage[T1]{fontenc}
\usepackage{graphicx}
\usepackage{lipsum} 
\usepackage{multirow} 
\usepackage{amsmath} 
\usepackage[capitalise]{cleveref} 
\begin{document}
\title{TSAK: Two-Stage Semantic-Aware Knowledge Distillation for Efficient Wearable Modality and Model Optimization in Manufacturing Lines}
\titlerunning{TSAK}
%
\author{Hymalai Bello\inst{1,2}\and
Daniel Geißler \inst{1} \and
Sungho Suh\inst{1,2}\and
Bo Zhou\inst{1,2}\and
Paul Lukowicz\inst{1,2}
}
\authorrunning{Bello et al.}
%
\institute{German Research Center for Artificial Intelligence (DFKI), Kaiserslautern, Germany\\
\and
Department of Computer Science, RPTU Kaiserslautern-Landau, Kaiserslautern, Germany
}
\maketitle              
\begin{abstract}
Smaller machine learning models, with less complex architectures and sensor inputs, can benefit wearable sensor-based human activity recognition (HAR) systems in many ways, from complexity and cost to battery life.
In the specific case of smart factories, optimizing human-robot collaboration hinges on the implementation of cutting-edge, human-centric AI systems. 
To this end, workers' activity recognition enables accurate quantification of performance metrics, improving efficiency holistically.
We present a two-stage semantic-aware knowledge distillation (KD) approach, TSAK, for efficient, privacy-aware, and wearable HAR in manufacturing lines, which reduces the input sensor modalities as well as the machine learning model size, while reaching similar recognition performance as a larger multi-modal and multi-positional teacher model.
The first stage incorporates a teacher classifier model encoding attention, causal, and combined representations.
The second stage encompasses a semantic classifier merging the three representations from the first stage.
To evaluate TSAK, we recorded a multi-modal dataset at a smart factory testbed with wearable and privacy-aware sensors (IMU and capacitive) located on both workers' hands.
In addition, we evaluated our approach on OpenPack, the only available open dataset mimicking the wearable sensor placements on both hands in the manufacturing HAR scenario.
We compared several KD strategies with different representations to regulate the training process of a smaller student model.
Compared to the larger teacher model, the student model takes fewer sensor channels from a single hand, has 79\% fewer parameters, runs 8.88 times faster, and requires 96.6\% less computing power (FLOPS).
Our results show that with TSAK distillation, the efficient model has significantly improved in recognition performance compared to the model trained without TSAK, with up to 10\% higher F1 score.

\end{abstract}
%
%
%
\section{Introduction}
\label{sec:Intro}
\begin{figure*}[!t]
    \centering
    \includegraphics[width=\columnwidth]{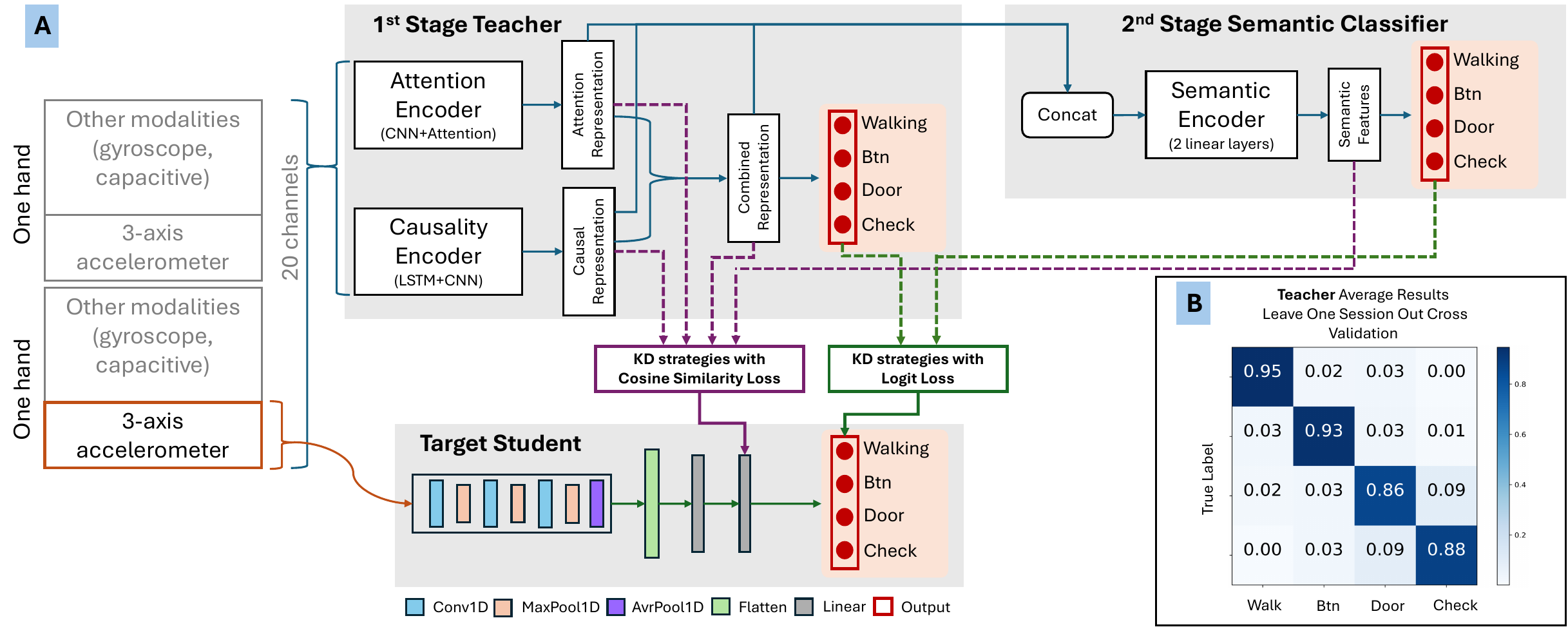}
    \vspace{-10pt}
    \caption{(A) In the TSAK knowledge distillation (KD) approach, five distillation methods were compared.
    The cosine similarity loss distills knowledge from one of the hidden vectors at a time; Attention Representation (Attn-Rep), Causal Representation (Causal-Rep), and Combined Representation (Combi-Rep).
    A shallow classifier is employed to merge and distill knowledge from all the hidden vectors simultaneously, preserving the semantics of the ground truth by logit-based KD (Semantic Classifier). Logit KD is also performed with the teacher's outputs.
    The output categories are walking, touching screen/buttons (Btn), opening/closing the door (Door), and working inside the factory module (Check).  
    (B) Teacher average results with an F1 score of 85.91\% across twelve users.}
    \label{fig:Dist}
    \vspace{-10pt}
\end{figure*}

The utilization of multi-modal sensing approaches, capturing the diversity of human behavior, has been widely leveraged to improve the accuracy and robustness of human activity recognition (HAR) systems.
The fusion of multimodal and multipositional information increases performance in the case of complementary sources, and it is robust to perturbations that may affect a particular sensing modality \cite{bello2023inmyface}.
However, most existing works prioritize improving the recognition performance while compromising efficiency, as smaller models with fewer sensor modality inputs usually suffer accuracy degradation.
Compared to other domains such as language and vision, wearable systems such as smartwatches or fitness trackers, are designed to be worn on the body for extended periods, making energy efficiency a critical factor. 
Prolonged battery life is essential to ensure user acceptance and adherence, as frequent recharging can be cumbersome and may lead to device abandonment. 
Furthermore, wearables often rely on small, low-power processors incapable of handling computationally intensive tasks.
However, most existing works leveraging multi-modal sensor fusion with machine learning techniques in pursuit of better accuracy, such as early or late fusion, transfer learning (TF), knowledge distillation (KD) \cite{gou2021knowledge,patidar2023vax,liang2022audioimu,bock2021improving}, and contrastive learning (CL) \cite{fortes2022learning,yoon2022img2imu}, have ignored efficiency aspects such as model compression, reduced complexity, and latency-awareness.

Although KD in particular has the potential to compress knowledge from a teacher model to a smaller learner model, studies, including ours, have shown that naïve KD with a small learner model has limited improvement compared to training the small model without KD, especially when the teacher model has more modalities and input channels.
An underlying contributing factor could be the semantic structural differences between the teacher and student making the representations from the teacher model less relevant for the student architecture.
We propose, TSAK, a Two-stage Semantic-Aware Knowledge distillation focusing on training efficient and fast student models with enhanced recognition performance, as shown in \cref{fig:Dist}.
In the first stage, we train a teacher model with both self-attention and LSTM branches extracting the attention and causal information.
In the second stage, the attention, causal, and combined representations are merged to feed into a semantic classifier.
The student model is a much smaller and more efficient model with simple convolution and linear operations, taking only 3 channels of sensor input (compared to up to 20 channels for the teacher).
To test the TSAK approach, we used the public dataset OpenPack \cite{yoshimura2024openpack} and our dataset at a smart factory testbed with smart gloves from 12 participants.
The \textbf{main contributions} of this work are described as follows: 
\begin{itemize}
    \item TSAK enhances a lightweight single-position accelerometer-only model (3-axis) for \textbf{wearable HAR in manufacturing lines} using two-stage KD from a larger multimodal and multipositional teacher model.
    \item Through ablation studies of different KD strategies including latent representations and logits \cite{hinton2015distilling,buciluǎ2006model,aguilar2020knowledge,guo2023semantic}, we have found out that using the second stage semantic classifier's logits output as the KD regularization improves the student performance significantly.
    \item 
    Experimental results with the OpenPack dataset and the dataset collected in this work show that TSAK increases the student F1 score by 5.4\%, and 10.5\%, respectively, with the student model being 79.0\% smaller, running 8.88 times faster, and 96.6\% less computation demanding (FLOPS).   
\end{itemize}
\vspace{-10pt}
\section{Related Work}
\label{sec:Related}
Representation learning has been extensively studied for HAR applications.
CL across inertial measurement unit (IMU) locations for wearable HAR was proposed in \cite{fortes2022learning} to learn a hidden representation from IMU sensors at different locations, but at the time of inference only use data from a single IMU.
Each IMU has its encoder and the objective is to guide the target encoder to learn from the best encoder, i.e., from the most informative IMU position. 
An improvement of F1 between 5\% to 13\% using PAMAP2 and Opportunity Dataset was obtained. 
This is an unimodal transfer learning strategy with no efficiency improvement in the target model.
In \cite{liang2022audioimu}, a KD method with multimodal fusion from audio to inertial sensors was proposed to transfer audio context information to the motion data for HAR, achieving an increase of 4.5\% F1 score for 23 activities (72.4\%). 
Their student employs acceleration and gyroscope data with separate branches of DeepConvLSTM \cite{bock2021improving} for each input type, leading to a student with 3.6 million parameters and high complexity due to the stack of LSTM layers.
VAX in \cite{patidar2023vax} is a cross-modality transfer learning where the video/audio (VA) is the teacher modality and X is a privacy-aware sensor from a selection of pervasive devices. 
Video/audio models are widely spread and trained with labeled data, alleviating the lack of training data with X sensors under the guidance of the VAX teacher.
An improvement of 5\% in accuracy compared to the baseline was obtained.

HAR with KD opens the possibility of transferring information from a cumbersome teacher model to a lightweight target-student model. 
A capable, smaller learner offers advantages, such as less memory, less power, and lower latency. 
KD has shown incredible potential in computer vision \cite{wang2021knowledge}.
The student can be guided by multiple methods, including logit-based, feature-based, and cross-modality among others \cite{gou2021knowledge}. 
However, most of these methods have not yet been evaluated in wearable datasets for HAR in manufacturing lines. 
\cref{tab:SOTA} reviews the state-of-the-art (SOTA) of engineering applications using cross-modality KD for wearable HAR. 
Overall, the SOTA shows that most solutions use complex structures for both the teacher and student network. 
For example, using deep convolutional LSTMs \cite{liang2022audioimu} and transformers \cite{liu2023emotionkd}.
These high complexity and high-performance networks are not yet supported by low memory, low power microcontrollers with limited FLOPS compared to a GPU \cite{nikolskiy2016floating}. 
In addition, the size difference between the teacher network and student network has to be moderate to increase the performance of the target model \cite{mirzadeh2020improved}. 
To the best of our knowledge, among the SOTA, there are no solutions with model enhancement through KD that produce target student models efficient enough for microcontroller deployment for wearables, which is one of the focuses of TSAK.

\begin{table}[!t]
\footnotesize
    \centering
    \caption{Engineering applications using cross-modality knowledge distillation for wearable HAR}  
    \vspace{-5pt}
    \begin{tabular}{c|c|c|c|c|c}
         Study & Accessory&Modality & Application& Improvement & KD Type \\
         \hline
         Liang \cite{liang2022audioimu} & Watch & Audio to (Acc+Gyro)& Daily Living &4.5\% F1 & Logit\\
         J Ni \cite{ni2022progressive} & Wristband & Skeleton to Acc& Body Motion&4.99\% Acc & Logit\\
         Liu \cite{liu2021semantics} & NA&Inertial to Videos & Body Motion&4.01\% Acc &Feat\\
         Ni \cite{ni2022cross}& Wristband& Videos to Acc& Body Motion&2.1\% Acc & Feat\\
         Liu. Y\cite{liu2023emotionkd} & NA& EEG to GSR* &Emotions&3.41\% F1 & Logit+Feat \\
         \textbf{Ours} &Gloves& (IMU+Cap) to Acc**&\textbf{Factory}&\textbf{5.4-10.5\% F1} & \cref{fig:Dist}\\
         \hline
    \end{tabular}
    {\centering *Electroencephalogram (EEG), Galvanic Skin Response (GSR). **Capacitive (Cap) \par}
  
    \label{tab:SOTA}
    \vspace{-10pt}
\end{table}
\addtolength{\tabcolsep}{1pt}


\section{Apparatus and Dataset}
\label{sec:HW}
\begin{figure}[!t]
    \centering
    \includegraphics[width=\columnwidth]{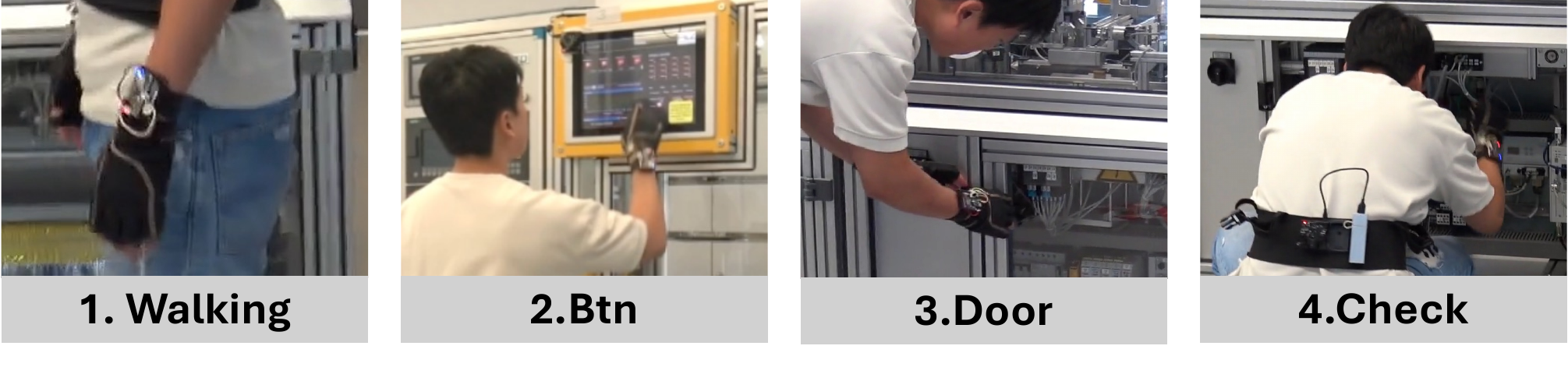}
    \vspace{-15pt}
    \caption{Activities dictionary in the smart factory testbed}
    \vspace{-10pt}
    \label{fig:Act}
    \vspace{-5pt}
\end{figure}

\textbf{Apparatus:} We use a glove-based system to monitor HAR in a smart factory environment \cite{bello2024besound}. 
The system backbone is the Adafruit Feather Sense development board with an Arm Cortex M4.
Two gloves are worn by the participants as shown in \cref{fig:Act}. 
Each glove has an IMU and four textile capacitive channels. 
Inside the Feather Sense, the BLE communication is handled by the Nordic nRF52832. 
The capacitive channels consist of four conductive thin patches distributed on the index, thumb, little finger, and around the wrist. 
Moreover, the IMU-selected placement is on the wrist.
This approach reduces the number of connections, and flexibility and comfort are considered. 
Noticeably, the gloves do not cover the entire area of the fingers, minimally affecting the user's mobility. 
Only the inertial (accelerometer and gyroscope) and capacitive data are used.
For a total of 10 channels per glove. 
Capacitive sensing has been used in textile designs such as neckbands \cite{cheng2013activity}, jackets \cite{bello2021mocapaci,bello2022move}, pants \cite{geissler2023moca} and particularly gloves \cite{bello2023captainglove}.
The textile capacitive sensor is based on the state-of-the-art capacitance-to-digital converter (CDC) FDC2214 following the design in \cite{bello2023captainglove}.
The excitation frequency of the CDC is set at 13.7 MHz with an 18uH external inductor and 33pf capacitor for each channel, operating with single-end sensing mode at a 50 Hz sampling rate.
Four channels of long electrodes (e-textile Shieldex Technik-tex P130+B) with the dimensions 0.55 mm wide and between 11-15 cm long were thermally bound from the sensing module to 3 fingers (the first, second, and fifth digits) and around the wrist. 

\textbf{Dataset:}
There are multiple datasets with wearable sensors, e.g. PAMAP2, Opportunity, WISDM, and RealWorld, among others \cite{huang2024survey}.
None of these human activity datasets meet the requirements to be representative and add completeness to the manufacturing line scenario. 
Moreover, publicly available sensor datasets in industrial settings are limited by difficulties in collecting realistic data, thus requiring close collaboration with industrial sites.
The Fraunhofer Institute has made publicly available\footnote{https://www.bigdata-ai.fraunhofer.de/s/datasets/index.html} a list of more than 120 open datasets from production environments, none of them related to human activity recognition and wearable devices.  
In HA4M \cite{cicirelli2022ha4m}, the first dataset about an assembly task with multiple vision sensors is introduced. 
The authors of HA-ViD \cite{zheng2024ha} and IndustReal \cite{schoonbeek2024industreal} also collected a human assembly video dataset, but still lack the wearable sensors for HAR. 
To address these challenges and contribute to research on HAR in industrial settings, OpenPack \cite{yoshimura2024openpack} recently introduced a dataset for packaging work recognition. 
In addition, in this work we have collected data from wearable sensors in a smart factory testbed, aiming to add completeness and relevance to our evaluation method. 
Our dataset selection criteria are based on three aspects: 1. HAR in manufacturing lines; 2. wearable, privacy-aware multimodal sensors; 3. sensor position on both hands.
This led us to collecting our own dataset, called Smart Factory Dataset, which is then complemented by the OpenPack dataset for secondary evaluation of our approach. 

\textbf{Smart Factory Dataset:}
Twelve volunteers were recruited.
They identify themselves as ten male and two female. 
Their age ranges from 23 to 59 years old (mean of 30.75). 
The height ranges from 160-184 centimeters (mean 178 cm). 
Only one of the participants was left-handed. 
The participants wore two gloves equipped with inertial and textile capacitive sensing. 
At the beginning of the experiment, the sensors' data is synchronized in front of a video camera. 
The camera time is then used as a global clock to synchronize the data from the two gloves. 
The volunteers were asked to walk around the factory modules and simulate working activities on the factory floor. 
\cref{fig:Act} depicts the activities performed by each volunteer were categorized as walking (1. Walk), touching the screen/buttons (2. Btn), opening/closing the door (3. Door), and working inside the factory module (4. Check). 
The activities were performed on each module (in total 6) in each session.
A total of five sessions per participant were recorded. 
In between every session, the hardware was removed from the wearer and a rest of ten to twenty minutes was enforced.
This makes the results accountable for the re-wearing of the system, which is typically expected in wearable devices. 
Each session lasted around 20 minutes on average.
One participant performed two sessions one day and three sessions another. 
One volunteer only performs three sessions in total. 
The participant with less than five sessions was only included in the training set.
For eleven participants, the data is split into 4 sessions for training and 1 session for testing. 
A 5-fold cross-validation with a leave-one-session-out evaluation scheme is performed. \footnote{All participants signed an agreement following the policies of the university's committee for protecting human subjects and following the Declaration of Helsinki \cite{Helsinki1975}. }

\textbf{OpenPack (Public):} 
The dataset contains packaging work activities in an industrial testbed, including work operations, actions, and outliers.
The main reasons for selecting OpenPack for analysis are the semi-realistic industrial setting and the configuration of sensors on volunteers' wrists, similar to our smart factory dataset.
It contains 53.8 hours of multimodal data, including key points, depth images, IMU data, and scarce readings from IoT devices (e.g., barcode scanners) \cite{yoshimura2024openpack}.
We focus on the IMUs data (accelerometer and gyroscope) from the users's wrists (right/left) and for the case of work operations activities.
The creators define the activities as 1. picking, 2. relocate item label, 3. assemble box, 4. insert items, 5. close box, 6. attach box label, 7. scan label, 8. attach shipping label, 9. put on back table, and 10. fill out.
Due to the low granularity of the labeling procedure, the categories are mixed within different labels. 
This led us to merge them into four classes; Pick (1,9,10), label (2,6,8), Assemble (3,4,5), and Scan.
The sampling rate for the sensors is around 30 Hz.
The idea is to transfer knowledge from a multimodal teacher to a student with one-handed acceleration data as input. 
Furthermore, we use five of eleven users' data to avoid faulty data. 
A 5-fold cross-user validation with a leave-one-session-out evaluation scheme is performed with five users' data.

\vspace{-10pt}
\section{Knowledge Distillation Approach}
\label{sec:TSAK}
\textbf{Pre-Processing Factory:}
The three accelerations ($m/s^2$) and the four capacitive channels are normalized between zero and one. 
The angular velocity channels are kept in their original range ($\pm$ 250 dps). 
This is followed by a 2-second resample window (100 samples at 50Hz). 
Then, a second-degree Butterworth low pass filter of 30 Hz was used to remove the jitter on the resampled signals. 
The resampling to 50Hz is for synchronization purposes with the video-ground truth with 50 frames per second. 
A sliding window of 2 seconds with a step size of 0.5 seconds is used.
After pre-processing, the dataset structure is ten channels for each glove with a window size of 2 seconds and 25\% overlapping. 
The ground truth of the worker's activity is extracted manually from the recorded videos. 

\textbf{Pre-Processing OpenPack:}
Acceleration and angular velocity data are resampled from 30 Hz to a 2-second resampling window (100 samples at 50 Hz) to match the factory dataset. 
The same 2-second sliding window and 25\% overlap are used. 
The dataset structure is six channels for each IMU on the user's wrists.

Next, we train the teacher for the knowledge distillation to the student. 
The Pytorch 2.1.0 version is used to train the neural networks (NN). 
The evaluation scheme is defined as a 5-fold cross-validation with leave-one-session out.  
The training of the NNs ran for 100 epochs with early stopping (patience 10) to avoid overfitting. 
And, a 64-batch size was selected.
The optimizer was Adadelta with a learning rate of 0.9. 
The loss function is categorical cross-entropy and the metric to monitor is accuracy. 

\textbf{Factory Teacher:} 
We have trained a multimodal and multipositional teacher model. 
Inertial and capacitive channels from both gloves (left and right hand) were fused, for an input size of 20 channels.
The teacher NN architecture details are in \cref{tab:NNDetails}. 
The structure contains two branches. 
One branch of the neural networks is focused on extracting features of the cross-channel interaction between the modalities and sensors' positions (\textbf{TASmart}), resulting in the hidden vector \textbf{Attn-Rep}. 
And, the second branch extracts the causality of the multimodal time series input (\textbf{TCSmart}), leading to the hidden vector \textbf{Causal-Rep}.
The causality extractor network is based on one-layer Long-Short-Term Memory (LSTM). 
Both networks are then concatenated (\textbf{Combi-Rep}) and fed into a classifier layer (Linear Layer), followed by an output layer with the softmax activation function. 
Combining the two concatenate NNs can capture spatial and temporal information, thus effectively solving complex time series problems. 
\cref{tab:Profile}, shows the profiling information of the teacher structure. 
These three hidden vectors are defined to compare the feature-based knowledge distillation (KD) method using different embedded representations within the teacher's NN structure. 
The feature-based KD is compared with logit-based KD. 
Furthermore, a shallow classifier merges and distills knowledge from the three hidden zones simultaneously, preserving the semantics of the ground truth by logit-based KD (see \cref{fig:Dist}).
\cref{fig:Dist}\textbf{B}, depicts the average results of the teacher for twelve volunteers and leave-one-session out cross-user validation scheme (F1 = 85.91\%). 

\textbf{OpenPack Teacher:}
The factory teacher is modified to match the dataset as follows: 1. the input structure only includes the acceleration and angular velocity channels and 2. the teacher structure is modified accordingly. 
The implementation details of the NNs are shown in \cref{tab:NNDetails}. 

\textbf{The Target Model:}
The idea is to explore the performance enhancement of small and simple networks that can be deployed in wearable and embedded devices with a reduced impact on power consumption and memory. 
The structure of the student/target NN is depicted in \cref{fig:Dist} (Bottom Left).
The input layer consists of one-handed acceleration data. 
The NN combines a feature extraction and a classifier with two linear layers. 
The NN is a CNN-based model for compatibility with supported operation on wearable embedded devices. 
To avoid degradation of the student performance the gap between the teacher and the student has to be moderate \cite{mirzadeh2020improved}.
The student model is 79\% smaller, 8.88 times faster, and 96.6\% less computation demanding (FLOPS) than the teacher’s, providing an embedded and sustainable solution.
The implementation details are in \cref{tab:NNDetails}.
The student's profile is compared with the teacher in \cref{tab:Profile}.

\textbf{TSAK Distillation Approach:}
The distillation approach is depicted in \cref{fig:Dist}. 
TSAK approach compares five distillation methods independently. 
The cosine similarity loss distills knowledge from one of the hidden vectors at a time; \textbf{Attn-Rep}, \textbf{Causal-Rep}, and \textbf{Combi-Rep}. 
Moreover, a shallow classifier merged and distilled knowledge from the three hidden vectors of the teacher simultaneously, preserving the semantics of the ground truth by logit-based KD (Semantic Classifier).
The semantic classifier is trained in an incremental stage. 
The frozen teacher is used to train the semantic classifier, which is subsequently also frozen and used to train the student. 
Logit-based KD is also performed with the teacher's soft outputs. 
The loss function $\mathcal{L}_{SC}$ for the logit-based (Logit) and semantic-based KD (SC-Logit) follows the \cref{eq:loss}.
Where $x$ is the input, $W$ are the student model parameters, $y$ is the ground truth label, $\mathcal{L}_{CE}$ is the cross-entropy loss function, $\mathcal{L}_{KL}$ is the KL-divergence loss, $\sigma$  is the softmax function parameterized by the temperature $T$ and $\alpha$ is a coefficient (weight). $z_s$ and $z_t$ are the logits of the student and teacher respectively.
In our experiments, $\alpha$ varies between 0.1, 0.5, 0.99, and 0.99.
$\tau$ values are 1, 4, and 20. 
This follows the experimental setting in \cite{cho2019efficacy}. 
\cref{fig:alphaT} depicts a comparison for different $\alpha$ and temperatures. 
For the case of feature-based KD, the loss function $\mathcal{L}_{F}$ follows the \cref{eq:lossF}, where $\mathcal{L}_{CSKD}$ is the cosine similarity loss, $h_t$ and $h_s$ are the hidden vectors of the teacher and student model, respectively. 
\begin{equation}
    \begin{split}
    \mathcal{L}_{SC}(x;W) &= \alpha\times\mathcal{L}_{CE}(y,\sigma(z_s; T=1)) \\
    &+ (1-\alpha)\times\mathcal{L}_{KL}(\sigma(z_t; T = \tau),\sigma(z_s, T = \tau))
    \end{split}
    \label{eq:loss}
\end{equation}
\vspace{-20pt}

\begin{equation}
    \mathcal{L}_{F}(x;W) = \alpha\times\mathcal{L}_{CE}(y,z_s) + (1-\alpha)\times\mathcal{L}_{CSKD}(h_t,h_s)
    \label{eq:lossF}
\end{equation}
\vspace{-10pt}

\begin{table}[!t]
    \centering
    \footnotesize
    \caption{Implementation details of the neural networks} 
    \begin{center}
    \begin{tabular}{c|c|c}
    Network & Layer & Details Kernel(K), Stride(S), Output(O)\\
    \hline 
    \multirow{6}{*}{TASmart*} & Conv1; MaxPool; Dropout& K=3, S=1, O=100; K=2, S=2; 0.2\\
                       & Conv1; MaxPool; Dropout& K=3, S=1, O=20; K=2, S=2; 0.2\\
                       & Conv1; MaxPool; Dropout& K=3, S=1, O=10; K=2, S=2; 0.2\\
                       & Self Attention1& (10,10)\\
                       & Self Attention2& (6,6)\\
                       \hline 
    \multirow{1}{*}{TCSmart**} & LSTM; Dropout& (10,10); 0.2\\
    \hline
    \multirow{6}{*}{TAPack***} & Conv1; MaxPool; Dropout& K=3, S=1, O=100; K=2, S=2; 0.1\\
                     & Conv1; MaxPool; Dropout& K=3, S=1, O=20; K=2, S=2; 0.1\\
                       & Conv1; MaxPool; Dropout& K=3, S=1, O=10; K=2, S=2; 0.1\\
                       & Self Attention1& (10,10)\\
                       & Self Attention2& (4,4)\\
                       \hline 
    \multirow{1}{*}{TCPack****} & LSTM; Dropout& (6,6); 0.1\\
    \hline
    \multirow{1}{*}{Classifier} & Linear; Linear; Linear& (120,10); (30,10); (10,4)\\

    \hline
    \multirow{4}{*}{Student} & Conv1; MaxPool; Dropout& K=3, S=1, O=100; K=2, S=2; 0.1\\
                       & Conv1; MaxPool; Dropout& K=3, S=1, O=5; K=2, S=2; 0.1\\
                       & Conv1; MaxPool; Dropout& K=3, S=1, O=5; K=2, S=2; 0.1\\
                       & Linear; Linear& (5,10); (10,4)\\
    \hline
    \end{tabular}
    {\centering \par *TASmart: Teacher Attention (TA) Branch for Smart Factory. TCSmart: Teacher Causality Extraction (TC) for Smart Factory. TAPack***: TA for OpenPack Dataset. TCPack****: TC for OpenPack\par}
    \end{center}
    \label{tab:NNDetails}
    
    \end{table}
    

\begin{table}[!t]
\vspace{-10pt}
    \centering
    \footnotesize
    \caption{Teacher and student profile comparison for Smart Factory and OpenPack dataset}
    \vspace{-5pt}
    \begin{tabular}{c|c|c|c|c}
         Model (Channels) & FLOPS & Latency & Throughput & Parameters \\
         \hline
         Teacher Factory (20) &651.85 M &27.91 ms &23.35 GFLOPS & 12.65K \\
         Teacher OpenPack (12) &418.08 M &18.68 ms &22.38 GFLOPS& 9.82K \\
         \textbf{Student (3)} & 22.48 M& 3.14 ms&7.15 GFLOPS&2.69K \\
         \hline         
    \end{tabular}

    {\centering Using Google Colab with a GPU V100 and DeepSpeed4Science \cite{song2023deepspeed4science} with Step = 5.\par}
    \vspace{-10pt}
    \label{tab:Profile}
\end{table}    

\section{Result and Discussion}
\label{sec:Results}
\begin{figure}[!t]
    \centering    
    \includegraphics[width=0.75\columnwidth]{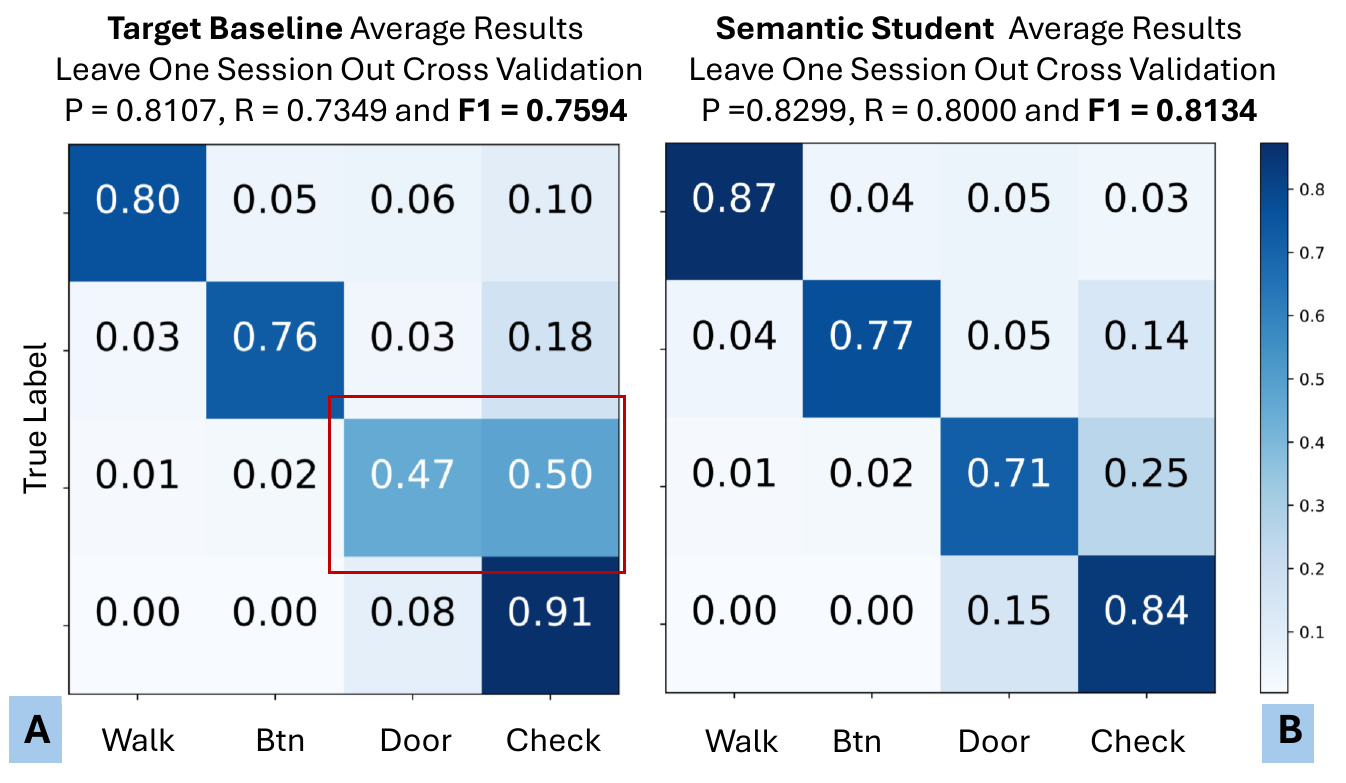}
    \vspace{-10pt}
    \caption{Twelve users' results in the smart factory scenario with right-handed acceleration channels as input. (A) Baseline; F1 of 75.94\% (B) Semantic student; F1 of 81.34\%}
    \label{fig:Results}
    
\end{figure}

\begin{figure}[!t]
    \centering   \includegraphics[width=0.75\columnwidth]{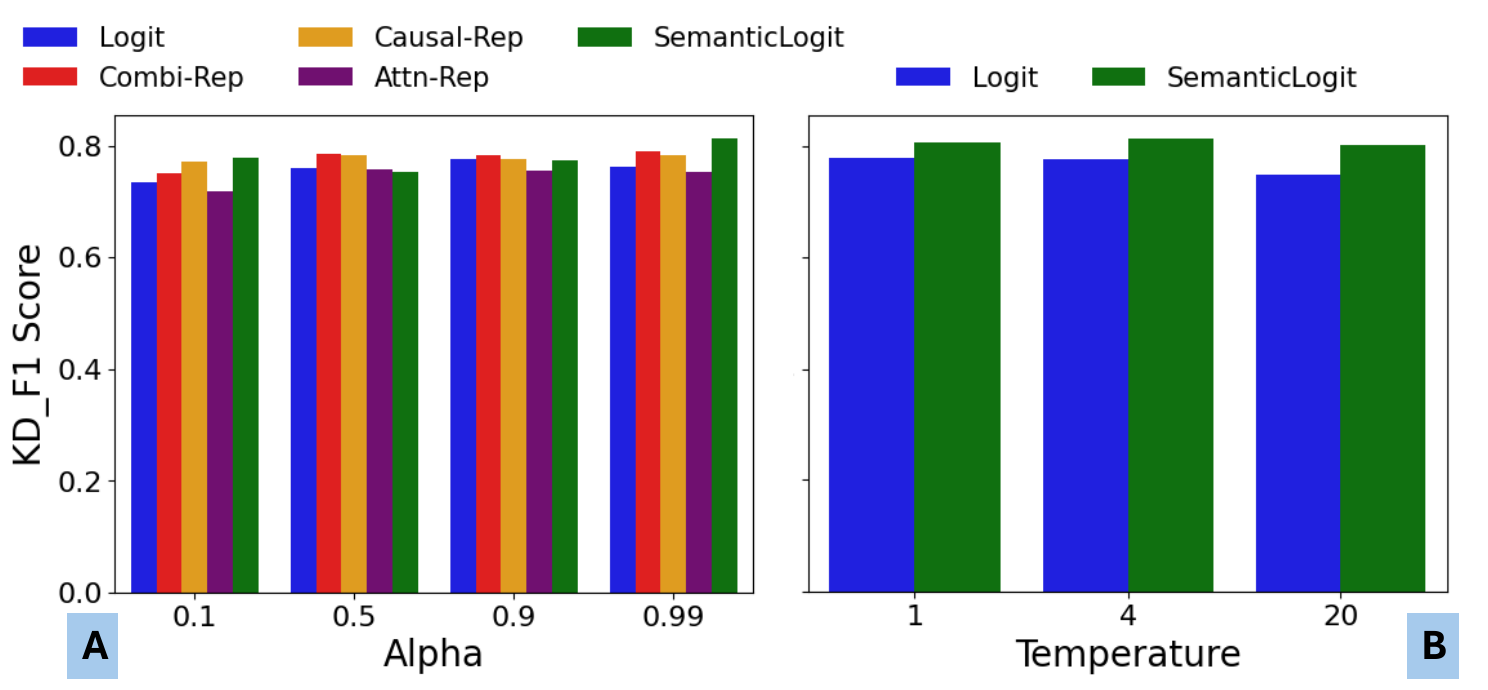}
    \vspace{-10pt}
    \caption{KD-based student comparison with right-handed acceleration inputs. (A) $\alpha$ variations. (B) Logit and SemanticLogit for different temperatures ($alpha=0.99$).}
    \label{fig:alphaT}
   \vspace{-10pt}
\end{figure}

\cref{fig:Dist}\textbf{B} shows the teacher results for the smart factory scenario. 
The model uses 20 inputs of inertial and capacitive channels from both gloves (F1 score of 85.91\%) with twelve users and four classes. 
The Door and Check classes have the higher confusion with 0.9\%. 
These two classes involve grabbing and pulling/pushing objects, in the first, the door is the object and in the second, the objects are inside the factory modules.
\cref{fig:Results}\textbf{A} shows the confusion matrix for the target baseline trained without any knowledge distillation (KD) technique and with the right-handed acceleration input data (3 channels) with 75.94\% F1 score. 
For the baseline, the classes Door and Check are 50\% confused.
Hence, our teacher contains complementary information compared to the target model. 
We have evaluated the performance with different $\alpha$ and temperatures as shown in \cref{fig:alphaT}. 
In general, the \textbf{Combi-Rep} performs best compared to the other hidden vectors. 
This vector contains causality and multimodal feature extraction information. 
The \textbf{Causal-Rep} is the second best in performance for different alpha values. 
This vector provides multimodal and multipositional causality knowledge to the student. 
The \textbf{Attn-Rep} (without causality) is the worst case compared to all the KD methods and for the right-hand target model without an increase in the F1 score compared to the baseline.
The increase in F1 score compared to the baseline is observed for $\alpha >0.5$ with the best $\alpha=0.99$. 

\begin{table}[!t]
    \footnotesize
    \caption{Results of KD approaches one-handed acceleration data for smart factory scenario} 
    \vspace{-5pt}
    
    \begin{center}

    \begin{tabular}{c|c|c|c|c}
    Function & Type&Precision & Recall & F1 Score\\
    \hline 
    \multirow{3}{*}{Teachers} &ATTN& 0.7703 & 0.8013 & 0.7840\\
                        & ATTN+LSTM& 0.8574 & 0.861 & 0.8591\\
                        & Semantic Classifier (SC) & 0.9060 & 0.9057 & 0.9058\\
    \hline 
    \multirow{8}{*}{\centering Right Hand Target}& Baseline& 0.8106 & 0.7349 & \textbf{0.7594}\\
                        & Logit& 0.7834 & 0.7545 & 0.7616\\
                        & Combi-Rep& 0.8209 & 0.7682 & 0.7887\\
                        & Causal-Rep& 0.8061 & 0.7652 & 0.7826\\
                        & Attn-Rep& 0.7835 & 0.7315 & 0.7528\\
                        & Merged Loss& 0.8029 & 0.7447 & 0.7676\\
                        & SC-Logit & 0.8299 & 0.8000 & \textbf{0.8134}\\
                        & SC-Feature & 0.8212 & 0.7568 & 0.7817\\
                        
    \hline 
    \multirow{8}{*}{\centering Left Hand Target}&Baseline&0.7512&0.6152&\textbf{0.6516}\\
                        & Logit&0.7240 &0.6510 & 0.6744\\
                        & Combi-Rep&0.7571&0.6770&0.7029\\
                        & Causal-Rep&0.7514&0.6633&0.6886\\
                        & Attn-Rep&0.7563&0.6674&0.6911\\
                        & Merged Loss & 0.7574 & 0.6612& 0.6885\\
                        & SC-Logit &0.7758&0.7055&\textbf{0.7283}\\
                        & SC-Feature & 0.7673 & 0.6792 & 0.7047\\                     

    \hline
    \end{tabular}
    \label{tab:ResultLeft}
    {\centering \par SC-Logit (Alpha 0.99 and Temperature 4) with F1-Score of 81.34\% and 72.83\%.\par}
    \end{center}
    \vspace{-10pt}
    \end{table}

The best result is obtained with the semantic classifier KD method (SC-Logit) with 81.34\% F1 score for an $\alpha = 0.99$ and $\tau=4$ (see \cref{fig:alphaT}\textbf{B}).
The semantic classifier has an F1 score of 90.58\% (+4.67\% $>$ teacher).
The confusion matrix of the semantic student trained with the SC-Logit KD method is presented in \cref{fig:Results}\textbf{B}.
The Door and Check classes reduce misclassification to 25\%, and classes such as Walk and interacting with buttons (Btn) have a 7\%, and 1\% increase in recall compared to the baseline (see \cref{fig:Results}\textbf{A}). 
\cref{tab:ResultLeft} compares the results between the teacher and different KD methods used to train a student with one-handed acceleration data as input. 
The teacher (CNN-LSTM) with two branches outperformed the teacher with a single-branch CNN, 85.91\% compared to 78.40\% F1 score.  
Based on $\alpha=0.99$, the best case, we have also compared the TSAK KD distillation method with a Merged-Loss (see \cref{tab:ResultLeft}). 
The Merged-Loss is based on the following equation; $\mathcal{L}(x;W) = \alpha\times\mathcal{L}_{CE}(y,z_s)+ (0.01)\times\mathcal{L}_{CSKD}(Attn-Rep,h_s) + (0.01)\times\mathcal{L}_{CSKD}(Causal-Rep,h_s) + (0.01)\times\mathcal{L}_{CSKD}(Combi-Rep,h_s)
$. 
Overall, the Merged-Loss outperformed the logit KD and the baseline. 
Moreover, in \cref{tab:ResultLeft}, we included a KD method based on the semantic classifier's last hidden vector (SC-Feature). 
SC-Feature KD distills knowledge to the one-handed student using the cosine similarity loss ($\alpha=0.99$).
The best result is from the student trained with the semantic classifier (SC-Logit) with an increase of 5.4\% and 7.67\% F1 score for the right-handed and left-handed targets, respectively.

\begin{figure}[!t]
\vspace{-10pt}
    \centering
    \includegraphics[width=0.75\columnwidth]{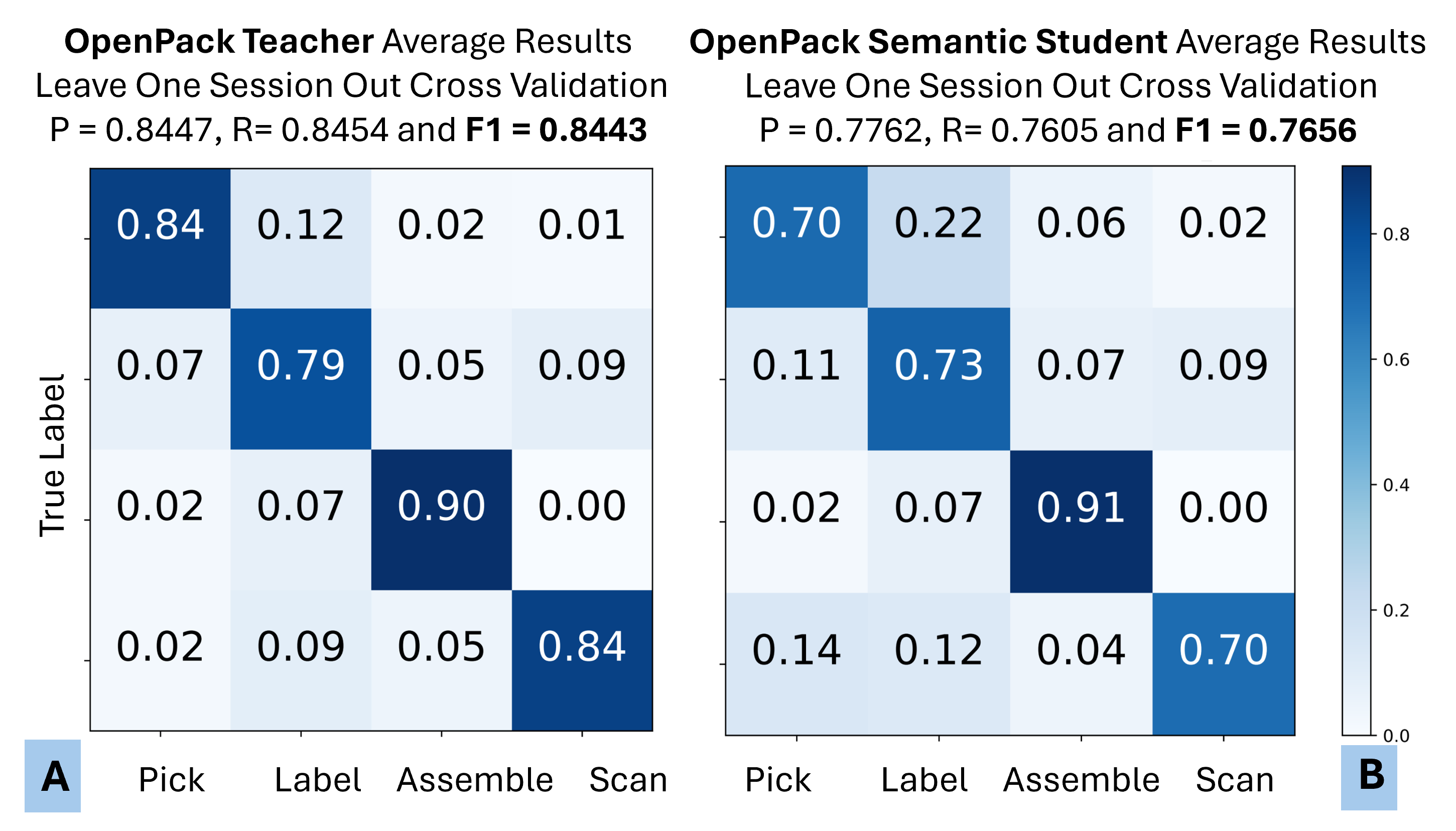}
    \vspace{-10pt}
    \caption{Five users' results with OpenPack. (A) Teacher results (12 channels); accelerometer and gyroscope (both hands). (B) Student results with right-handed accelerations as input (Semantic KD) and 10.5\% F1 score increase.}
    \label{fig:ResultsOpen}
    \vspace{-15pt}
\end{figure}

\cref{fig:ResultsOpen}\textbf{A} presents the results for the OpenPack teacher with twelve input channels and  84.43\% F1 score. 
These results are for five participants with a leave-one-session cross-user validation scheme (5 sessions in total). 
\cref{fig:ResultsOpen}\textbf{B} shows the best-distilled student for the OpenPack (SC-Logit KD) with an increase of 10.5\% in F1 score compared to the baseline (the target model trained without KD). 
The semantic classifier has an F1 score of 86.16\% (1.73\% $>$ teacher).

\cref{tab:ResultOpen} compares the results of the teacher, baseline, and the right-handed student with acceleration channels as input and trained with the five types of KD.  
The training method for the teacher and the baseline NNs is the same as in the factory dataset case. 
And, to train the different distilled students the settings are the best cases from the factory dataset. 
This means these results do not include hyperparameter tuning or any optimization method.
We can expect that tuning the NNs for this specific dataset will increase the performance in the future.
For both datasets, the second best KD method was the feature-based with the \textbf{Combi-Rep}. 
Overall, our results show the potential of cross-modal knowledge distillation for inertial sensing systems for HAR in a factory testbed. 

The majority of distilled student models (excluding \textbf{Attn-Rep}) with $\alpha>0.5$ outperformed the target model baselines. 
This increase in performance indicates that the multimodal and multipositional teacher effectively guides the student.
The semantic classifier (SC-Logit) as the teacher is the most effective KD method for both datasets. 
Importantly, the semantic classifier is trained in an incremental step from the frozen teacher. 
In addition, SC-Logit combines feature-based knowledge with logit-based knowledge.
This could be the reason for performance improvement. 
With this method, the lightweight single-position (3-axis) accelerometer-only model is compressed and improved. 
This means that our solution retains the practicability of using only three channels IMU available on even the simplest smartwatches, and at the same time, improves performance.
For a student model 79\% smaller, 8.88 times faster, and 96.6\% less computation demanding than the teacher's. 
On the other hand, our solution has \textbf{limitations} and possibilities for improvement: 

\begin{enumerate}
    \vspace{-2mm}
    \item We employ the \textbf{Kullback-Leibler Divergence loss} successfully controlling the ``soft" targets via the temperature scaling parameter. 
    In \cite{kim2021comparing}, the authors proved that the MSE loss outperforms the KLD loss, explained by the difference in the penultimate layer representations between the two losses.
    \item We \textbf{manually varied the $\alpha$ and $T$} of the KD. 
    This can be substituted for an automatic parameter search \cite{li2023automated}. 

    \item In the future, \textbf{quantization-aware KD} will be a strong method to evaluate solutions for wearable HAR. 
    The idea is to coordinate the quantization and KD approach to fine-tune a quantized low-precision student network \cite{kim2019qkd} for an optimized embedded solution.
    \item We evaluated our method on \textbf{two wearable HAR datasets in manufacturing lines}. 
    This is mainly due to the restricted availability of open datasets with wearable sensors for HAR in industrial settings. 
    However, we believe the method can easily be applied to other scenarios of HAR.
    \vspace{-3mm}
\end{enumerate}




\begin{table}[!t]
\footnotesize
    \centering
    \caption{Results of KD approaches with right-handed acceleration data for OpenPack}
    \begin{tabular}{c|c|c|c}
         KD Type & Precision & Recall& F1 score\\
         \hline
         Teacher (CNN-LSTM) &0.8447 &0.8454 &0.8443\\
         Semantic Classifier &0.8657&0.8596& 0.8616\\
         Baseline&0.6799 &0.6541 &\textbf{0.6606}\\
         Logit&0.7119 &0.7026 &0.7060 \\
         Combi-Rep& 0.7654&0.7567 &0.7571\\
         Causal-Rep& 0.7292& 0.7329&0.7306 \\
         Attn-Rep& 0.7493&0.7402 &0.7422 \\
         Merged-Loss&0.7608 &0.7478 &0.7531 \\
         SC-Logit&0.7762 &0.7605 &\textbf{0.7656} \\
         SC-Feature&0.7627 &0.7582 &0.7598\\
         \hline
    \end{tabular}
    \label{tab:ResultOpen}
    \vspace{-10pt}
\end{table}

\vspace{-10pt}
\section{Conclusion}
\label{sec:Conclusion}

In this paper, we presented TSAK, prioritizing both model efficiency and precision for activity recognition, a two-stage semantic-aware knowledge distillation approach.
In the industrial manufacturing scenario, we tested TSAK with two datasets with smart wearables of wrist and hand-worn sensors: OpenPack and our own recorded dataset at a smart factory testbed.
We show that the second stage contributes significantly to the lightweight student model without altering its architecture.
The student model takes only 3-axis accelerometer data and has only simple 1D convolution and linear operations with 2.69k parameters, which is qualified to be deployed on most modern microprocessors.
The student model enhanced with TSAK has up to 10\% better F1 score compared to the same model without KD.
With merely 3.4\% the computational demand of the first stage teacher model, the student model's F1 score can reach 4.57\% short of the teacher model in the best scenario.
By leveraging the strengths of multi-modal sensing and machine learning techniques, while prioritizing energy efficiency and model compactness, we have demonstrated significant improvements in recognition performance, computational speed, and energy consumption. 
As we look to the future, TSAK underpins the development of sustainable, efficient, and accurate wearable HAR systems that can be seamlessly integrated into a variety of applications.

\vspace{-10pt}
\subsubsection{Acknowledgements} The research reported in this paper was partially supported by the German Federal Ministry of Education and Research (BMBF) in the project VidGenSense (01IW21003) and by the European Union in the project SustainML (101070408).

%
%
%
%
\bibliographystyle{splncs04}
%
\vspace{-10pt}
\bibliography{References}
\end{document}